\documentclass[journal]{IEEEtran}

\usepackage[normalem]{ulem}
\usepackage{gensymb}
\usepackage[export]{adjustbox}
\usepackage{float}
\usepackage{svg}
\usepackage{xcolor}
\usepackage{url}
\usepackage{amsmath}
\usepackage{subcaption}
\usepackage{tikz}

\renewcommand\fbox{\fcolorbox{red}{white}}
\newcommand\submittedtext{%
	\footnotesize This work has been submitted to the IEEE for possible publication. Copyright may be transferred without notice, after which this version may no longer be accessible.}

\newcommand\submittednotice{%
	\begin{tikzpicture}[remember picture,overlay]
		\node[anchor=south,yshift=10pt] at (current page.south) {\fbox{\parbox{\dimexpr0.65\textwidth-\fboxsep-\fboxrule\relax}{\submittedtext}}};
	\end{tikzpicture}%
}

\begin{document}

\title{Acoustic tactile sensing for mobile robot wheels}

\author{Wilfred~Mason,
        David~Brenken,
        Falcon~Z.~Dai, 
        Ricardo Gonzalo Cruz Castillo,
        Olivier~St-Martin~Cormier, 
        and~Audrey~Sedal
        
\thanks{W. Mason, D. Brenken and O. St-Martin Cormier are with McGill University, Montreal, Canada.}
\thanks{A. Sedal is with McGill University and Mila - Quebec AI Institute.}
\thanks{F. Z. Dai was with TTI-Chicago, Chicago, USA.}}

\maketitle

\submittednotice

\begin{abstract}
Tactile sensing in mobile robots remains under-explored, mainly due to challenges related to sensor integration and the complexities of distributed sensing. In this work, we present a tactile sensing architecture for mobile robots based on wheel-mounted acoustic waveguides. Our sensor architecture enables tactile sensing along the entire circumference of a wheel with a single active component: an off-the-shelf acoustic rangefinder. We present findings showing that our sensor, mounted on the wheel of a mobile robot, is capable of discriminating between different terrains, detecting and classifying obstacles with different geometries, and performing collision detection via contact localization. We also present a comparison between our sensor and sensors traditionally used in mobile robots, and point to the potential for sensor fusion approaches that leverage the unique capabilities of our tactile sensing architecture. Our findings demonstrate that autonomous mobile robots can further leverage our sensor architecture for diverse mapping tasks requiring knowledge of terrain material, surface topology, and underlying structure. 
\end{abstract}


\section{Introduction}

This work presents an acoustic tactile sensor that measures deformation around the circumference of a mobile robot's wheel. Tactile sensing could be a useful modality for navigation and path planning of mobile robots in contact-rich, dark, occluded, or otherwise challenging environments. Contact information including interactions with terrain, existence and shape of small obstacles on the ground, and collision information could all be encoded through tactile sensors. Such information allows mobile robots to adapt their locomotion style under new terrain, map features and materials in the environment, and fine-grained identification of collisions.

Presently, mobile wheeled robots mainly obtain information about terrain and obstacles through exteroceptive sensors such as cameras and LiDAR \cite{marti2019review}. In contrast, mobile legged robots have benefited from tactile sensors placed in their feet to improve precision and balance \cite{stone2020walking}, while manipulators have demonstrated dexterous behaviours by leveraging tactile sensor information \cite{hogan2022finger,dongimproved}. Robotic components which  have large distributed surfaces or complex motion suffer a morphology challenge that makes tactile perception difficult to integrate. Wheels in particular are constantly rotating, compliant in contact with the environment, and do not have locations to easily mount exteroceptive sensors for wide views. Existing tactile sensing solutions for wheels, tires, and other continuously deformable robot structures include pressure sensor arrays \cite{chen2021deepening}, expensive optical technologies such as Bragg diffraction gratings \cite{lilge2022continuum} or lower-cost optical systems \cite{zhao2016optoelectronically,to2018soft}. Many works demonstrate usefulness of tactile sensors for deformable systems in contact-rich settings. Yet, most existing tactile sensors require arrays of multiple input/output points, use brittle fibers that are not resilient to external forces and vibrations, or lose significant information when bending of the robot structure occludes a camera view. In spite of the high potential shown by tactile sensing systems, design and implementation of novel tactile sensors for mobile robots remains under-explored.

This paper examines an acoustic solution which has potential to provide rich tactile data while addressing the aformentioned challenges of shape, motion, deformability, and cost. We propose a sensor that couples an acoustic rangefinder with a flexible waveguide (tube) wrapped around the wheel of a mobile robot so as to act jointly as a tactile sensor and tire.
Acoustic rangefinders are widely used due to their simplicity and long-distance range.
Here, deflection of the tire under terrain or obstacle contact, and acoustic phenomena such as absorption and reflection, provide information contact information to the robot. Only one acoustic rangefinder at the tube's end is needed for perception around the entire wheel circumference.
This work includes the design and construction of a prototype acoustic-tactile tire sensor and its integration into a mobile robot. The paper further demonstrates its capabilities in terrain classification, shape classification of small obstacles, and contact localization. The main contributions are:

\begin{enumerate}
    \item Design of a low-cost and easy-to-deploy acoustic sensing architecture  for mobile robot wheels that relies on a single off-the-shelf rangefinder and a deformable waveguide,
    \item Data-driven classifiers for five terrain types and two obstacle shapes with varying size,
    \item A simple heuristic based on first principles for contact localization,
    \item A suite of physical experiments evaluating the classifiers and heuristic with a comparison between acoustic data and data from an inertial measurement unit (IMU).
\end{enumerate}

Together, these contributions show that acoustic tactile tire sensors overcome challenges related to morphology and occlusion to viably provide important, granular tactile data to mobile robots. Section \ref{sec:related-work} describes related work forming the basis for this contribution. Section \ref{sec:sensor} describes the physical basis for the sensor and the integrated mobile robot prototype. Section \ref{sec:expts} describes the physical experiments, data processing, and task evaluation frameworks for the two classification tasks and heuristic. Section \ref{sec:results} gives the experimental results and Section \ref{sec:discussion} analyzes the impact and limitations of the results. Section \ref{sec:conclusion} summarizes the work and future directions.

\begin{figure*}
    \centering
    \begin{subfigure}{0.345\textwidth}
        \centering
        \includegraphics[width=0.88\textwidth]{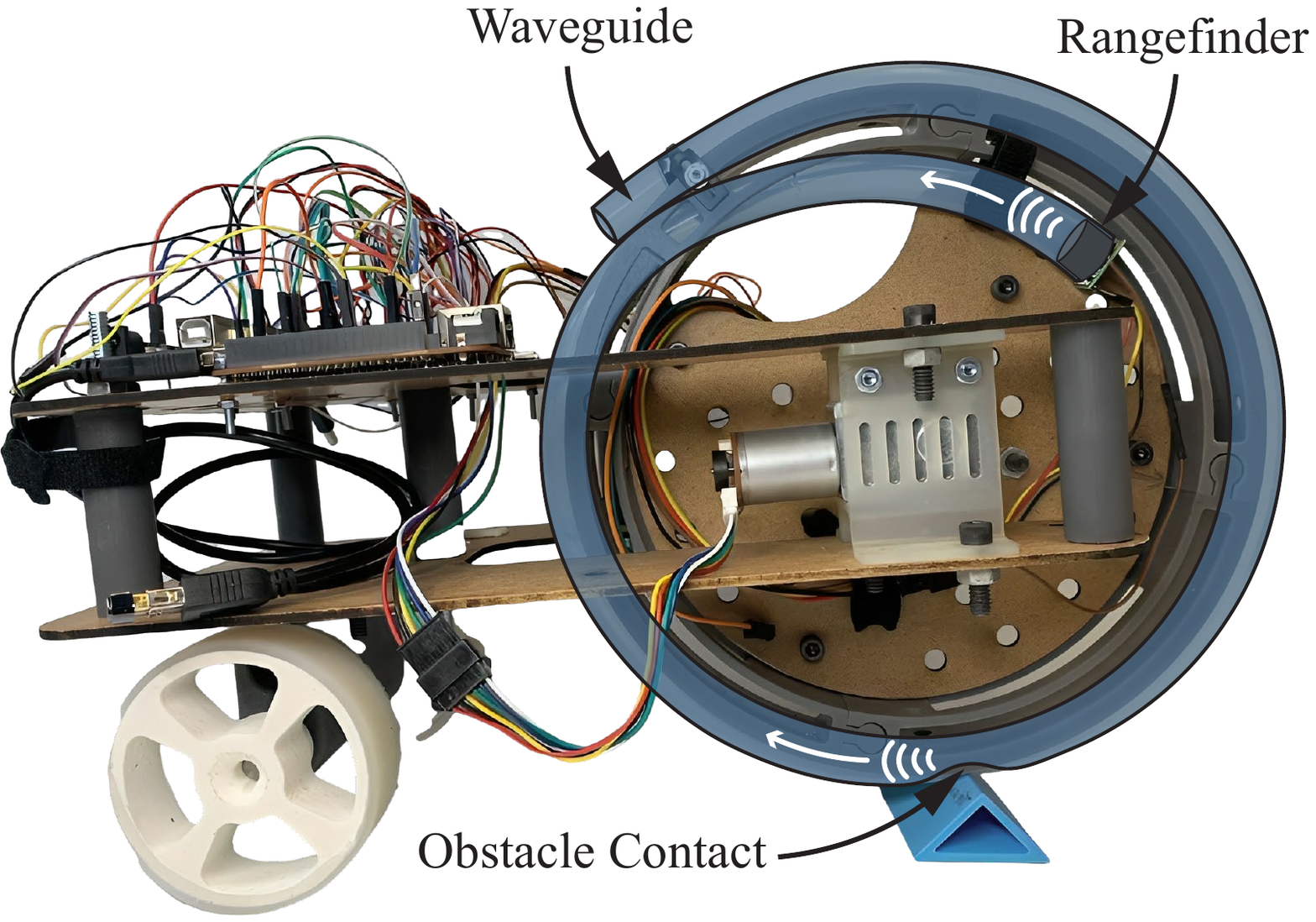}
        \caption{Mobile robot with wheel sensorized by an acoustic waveguide (tube) wrapped around its circumference and an acoustic rangefinder. Example sent/received acoustic pulses in white.}
        \vspace{2pt}
        \label{sfig:mobile_robot}
    \end{subfigure}\hfill
    ~    
    \begin{subfigure}{0.315\textwidth}
        \centering
        \includegraphics[height=0.5in]{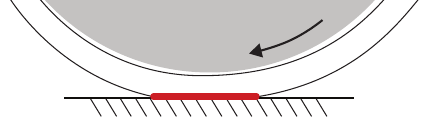}
        \caption{Tube indentation on flat surface.}
        \includegraphics[height=0.5in]{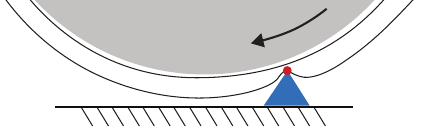}
        \caption{Tube indentation on obstacle.}
        \includegraphics[height=0.5in]{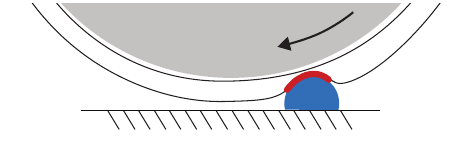}
        \caption{Tube indentation on obstacle.}

    \end{subfigure}\hfill
    ~
    \begin{subfigure}{0.315\textwidth}
        \centering
        \includegraphics[height=0.5in]{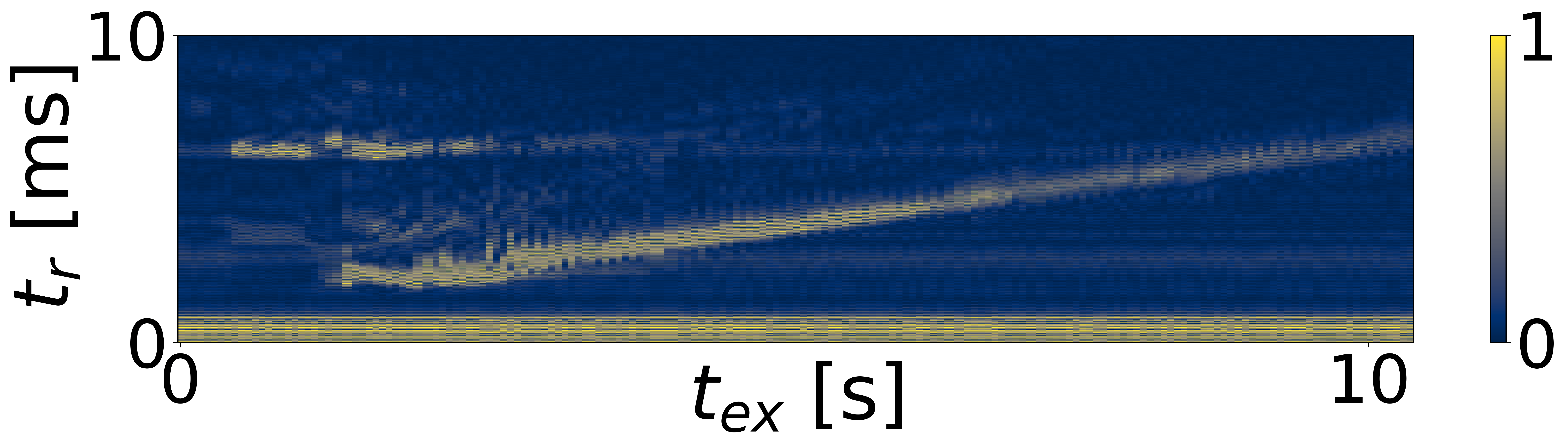}
        \label{fig:flat_trace}
        \caption{Trace history from flat ground contact.}
        \includegraphics[height=0.5in]{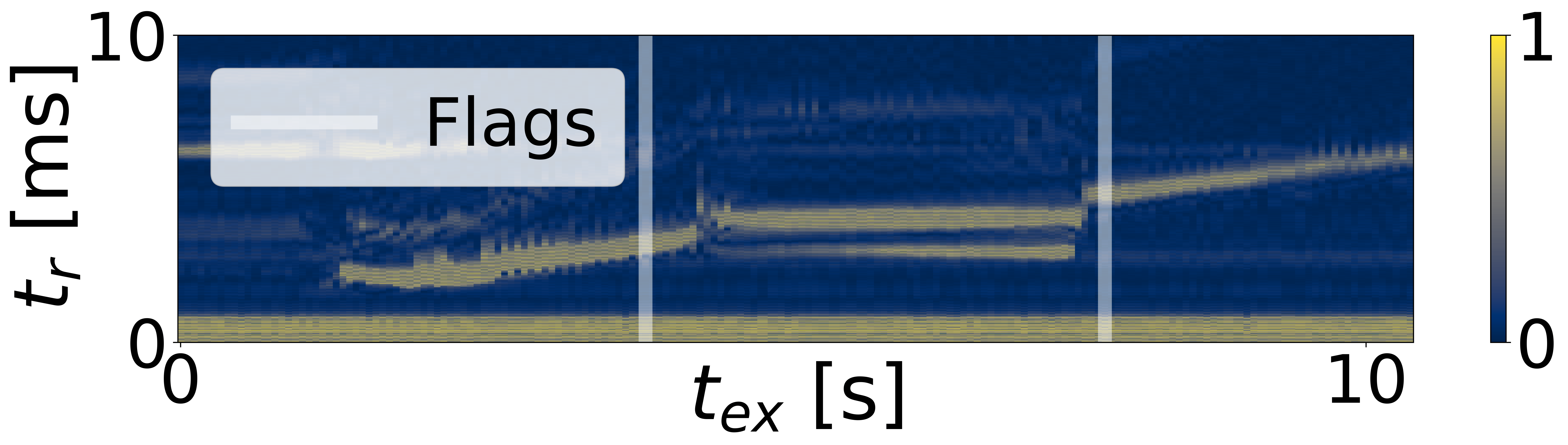}
        \label{fig:tri_trace}
        \caption{Traces from triangular obstacle.}
        \includegraphics[height=0.5in]{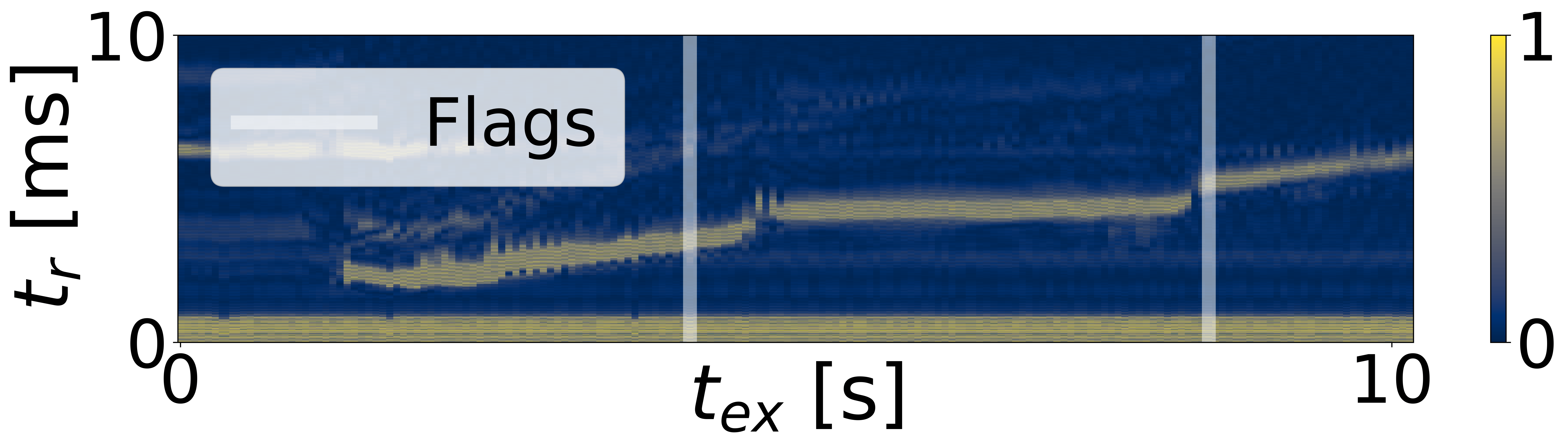}
        \label{fig:round_trace}
        \caption{Traces from round obstacle.}
    \end{subfigure}\hfill
    
    \caption{Overview of system design and sample data from obstacle contact.}
    \label{fig:composite_fig}
\end{figure*}

\section{Related Work}
\label{sec:related-work}

Previous work has shown that tactile information is valuable for 3D object mapping \cite{suresh2021tactile}, terrain classification \cite{guo2020soft,ding2022pressing,hoepflinger2010haptic}, manipulation and navigation in cluttered or contact-rich environments \cite{albini2021exploiting, thomasson2022going}, slip detection \cite{marchetti2020barefoot,liu2022design}, state estimation and stability analysis \cite{edlinger2021stability,mardani2017simultaneous,ichimura20163d}. Such information can be integrated into a mobile robot's controller or policy to improve its performance \cite{tomioka2022autonomous}.

In addition to force or pressure transceiver arrays \cite{nagatani2009accurate,ding2022pressing,wu2019tactile}, a variety of new tactile sensor architectures have emerged that use single-point sensors with specially designed enclosures that guide the input to the single-point sensor. Visuo-tactile sensors such as Gelsight \cite{dongimproved}, STS \cite{hogan2022finger} and others \cite{van2020large} use photometric stereo and/or point tracking to reconstruct the geometry of an object from its contact deformations with a reflective or marker-covered membrane. Such sensors have enabled in-hand manipulation with smaller-scale object information \cite{pan2023hand} and larger-scale proprioception through contact detection along the length of a robot arm \cite{van2020large,luu2023simulation,le2023viart}. Such sensors further measure fine detail and give information about contact locations and forces. Mounted to the feet of a walking robot, they have allowed locomotion in challenging environments \cite{stone2020walking,zhang2021tactile}. Yet, in all of these works light is required to travel directly between the deformable membrane and the camera. Occlusions due to specific membrane deformations or robot configurations that block the path of light to the internal camera would reduce their capabilities. Other optical tactile sensors leverage light attenuation through bending. Soft optical waveguides which deflect optical beams under bending have been used to map surfaces \cite{zhao2016optoelectronically}. 

Acoustic sensing frameworks have been proposed that offer the same structural advantage of visuo-tactile sensing with a different possible range. Acoustic vibrations can travel through elements that partially or fully occlude light \cite{lin2022wearable,dwivedi2018advances}. They further offer new scalability options arising from the spectrum of acoustic wavelengths. Inspired by key methods such as sonar which is frequently used by robots low-light settings such as underwater \cite{jang2021multi}, researchers have combined acoustic sensing with external structure for passive and active applications. 
Exciting a rigid robot structure with piezoelectric speakers and contact microphones can localize contact on rigid robots \cite{fan2022enabling}. Acoustic tactile sensors further have well-demonstrated functionality in highly deformable structures. Echotube \cite{tejada2019echotube} combines an acoustic rangefinder with a tubular waveguide to locate contact along the length of the tube. Other research places the sender and receiver at opposite ends of the waveguide to achieve strain, twist, and bend sensing \cite{chossat2020soft}. Active acoustic sensing has enabled proprioception \cite{zoller2020active,randika2021estimating} and texture sensing \cite{lu2023active} for soft robots through antagonistic placement of a microphone and chirping speaker or oscillator within the cavity. Passive acoustic sensing through contact microphones has enabled contact detection for soft robotic grippers \cite{mikogai2020contact}. Other passive sensors leverage acoustic resonance to measure force, pose, and contact in soft cavities embedded within soft robots \cite{li2022resonant}. Recent work presents a detailed comparison of passive and active acoustic sensing paradigms \cite{wall2023passive}, clarifying the advantages of ambient sounds and active reflectance-based measurements for proprioception and characterization of contact. 

Wheeled mobile robots could benefit from the capabilities given by tactile sensing to assist in navigation, mapping, state estimation and other tasks. However, integration of tactile sensors into wheels is challenging due to the large distributed area of the wheel, the wheel's circumference which naturally causes occlusions, and continuous rotation. Existing tactile solutions for mobile robots often rely on single-point probes \cite{giguere2011simple,kannan2020material} or arrays of many sensors such as inductive pressure pads \cite{chen2021deepening, marchetti2020barefoot} and infrared \cite{lauria2002octopus}. A transparent wheel \cite{yao2023wheel} has been developed with an internal camera capable of visualizing terrains to estimate surface geometry, wheel-terrain-interaction parameters, and soil flow. Yet, this camera does not provide a view around the wheel's circumference. Designing visuo-tactile wheels with a barrel shape \cite{le2023viart} and mounting a camera along the axel provides one way to overcome these morphological challenges. Yet, in many mobile robots and vehciles, the relative diameter of the tire in comparison with the wheel diameter would cause highly occluded views at the axel. Combining acoustic sensing with a flexible external waveguide into the structure of a wheel might offer the possibility for rich tactile sensing across the large distributed tire area.

\section{Sensor Description and Integration}
\label{sec:sensor}
We contribute a sensor which combines an acoustic sensor and a tubular waveguide wrapped around the wheel of a mobile robot. Throughout rotation of the robot's wheel, received acoustic reflections allow the mobile robot to detect locations of contact with the environment, enabling classification of ground texture and features. 

\subsection{Sensor}

The active component of the sensor is an acoustic rangefinder. The rangefinder emits a ranging pulse that travels through a tube wrapped around the wheel of a mobile robot. The tube acts as a waveguide, causing the ranging pulse to travel the entire circumference of the wheel. As the wheel rolls across different terrains, the waveguide is indented from ground and obstacle contact. The send pulse reflects off indentations along the waveguide and returns to the rangefinder. The rangefinder contains a microphone that records the amplitude of the incoming acoustic waves. The time differences between the transmission of the send pulse and the acquisition of high amplitude reflected waves, as well as other acoustic features including intensity or higher-order effects, might be used to localize and characterize deformations along the wheel. A schematic of the sensing system is shown in Figure \ref{sfig:mobile_robot}; travel and reflect of the sound wave is depicted. The following physical relationships characterize trade-offs between sensor speed, resolution, and possible contact area coverage.

The minimum query time ($\Delta t_\mathrm{range}$) to the acoustic sensor (send and return pulse) is determined by the length of the waveguide ($L$) and the speed of sound ($c$) in the waveguide medium.

\begin{equation}
    \Delta t_{\mathrm{range}} = 2L/c
    \label{eq:query}
\end{equation}

The waveguide's total length can be separated into two lengths, the part contained within the wheel hub ($L_{\mathrm{inner}}$) and the part that wraps around the wheel ($L_{\mathrm{outer}}$). Relating the length of the waveguide to the wheel radius ($r_{\mathrm{wheel}}$), and considering the wheel's angular speed ($\omega$) the number of ranging cycles ($\mathrm{C}$) acquired per wheel rotation ($\mathrm{R}$) is found. This is a useful metric that estimates the information density of the sensor by relating query time (Eqn. \ref{eq:query}) to the angular speed of the mobile robot.

\begin{equation}
    \mathrm{C/R} = \frac{2\pi}{\omega}\frac{1}{\Delta t_{\mathrm{range}}} = \frac{\pi c}{\omega(2\pi r_{\mathrm{wheel}} + L_{\mathrm{inner}})}
\end{equation}

The smallest distance ($d_{\mathrm{min}}$) that two indentations along the waveguide can be separated from each other and still be independently localized is given by the speed of sound in the waveguide medium and the width of the send pulse \cite{maier2018medical}. The width of the send pulse is related to the wavelength of the acoustic wave ($\lambda$) and the number of cycles ($n_{\mathrm{cycles}}$) in the send-pulse.

\begin{equation}
    d_{\mathrm{min}} = \frac{n_{\mathrm{cycles}}\lambda}{2}
\end{equation}

Following these physical principles, we built a tactile sensor prototype. The waveguide was a silicone rubber tube with a circular cross-section, 5/8" ID, 3/4" OD, Durometer 50A (McMaster-Carr, 51135K45). The acoustic rangefinder was a 42~kHz off-the-shelf model (MaxBotix, MB1010 LV-MaxSonar-EZ1) with a maximum ranging distance of 6.45~m. We directly mechanically coupled the rangefinder to the waveguide, and wrapped the waveguide around a wheel of 27~cm diameter without tube buckling. The rangefinder and a portion of the waveguide (length $\approx 15~\mathrm{cm}$) corresponding to the ``dead zone'' (i.e., manufacturer-specified minimum possible ranging distance) of the rangefinder were contained within the wheel hub. The other end of the waveguide was coupled to the wheel hub using a 3D-printed adapter and left open. This sensor design allows the rangefinder to be protected from the environment without impeding flexibility in the selection and integration of different waveguides. 

We acquired data from the sensor through the following steps. At 20~Hz, a microcomputer (BeagleBone Black) sent a trigger to the rangefinder which then emitted an acoustic pulse of 42~kHz. A microphone within the rangefinder measured acoustic amplitude. The microphone outputted analog output data (0-1.8~V) to an analog-to-digital converter (ADC) at 200~kHz. One ranging cycle spanned from a send pulse trigger event until 4000, 16-bit ADC samples were recorded. A buffer the size of one ranging cycle was stored on the BeagleBone. Ranging cycle data was subseqently sent to a host computer using sockets. Figures \ref{fig:composite_fig}b-g show possible indentation states for the tire and examples of corresponding acoustic data with differing reflection features.

\subsection{Integration into Mobile Robot}

We mechanically and electronically integrated the tactile sensor into a mobile robot. The mobile robot we designed is shown in Figure~\ref{sfig:mobile_robot}. The mobile robot is a three-wheel rover with a sensorized, leading drive wheel and passive rear wheels. The leading wheel is motorized by a DC gear motor (Walfront via Amazon.ca). A magnetic encoder coupled to the input shaft of the gear motor is used to record the position of the wheel. The leading wheel consists of a custom rim wheel that can accommodate waveguides with different diameters. The rim contains an exit hole for the waveguide which emanates from within the wheel hub, and a mounting feature to secure the end of the waveguide. The center of the wheel and robot chassis are structured with laser cut fibreboard, which house the DC motor and electronics. Data and power lines from the embedded acoustic sensor run within a through-hole type, slip-ring connector, and then connect to the onboard electronics. The waveguide diameter, hardness and wheel diameter were chosen empirically to avoid buckling of the waveguide when wrapped around the rim while allowing deformation of the waveguide under contact.

\begin{figure}
    \centering
    \includegraphics[width=\linewidth]{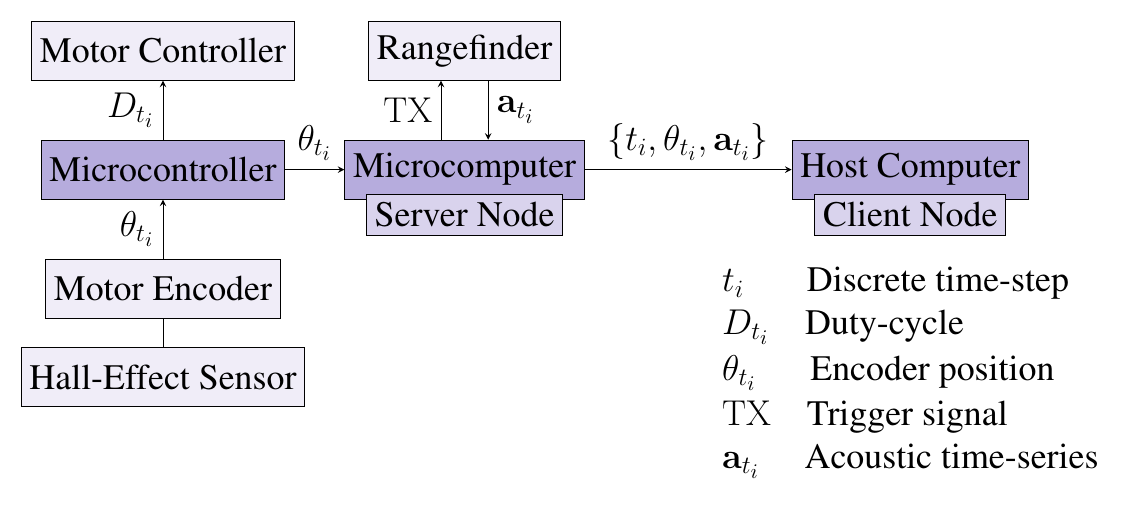}
    \caption{Operational diagram for the system including motor control, sensor data acquisition from an encoder and an acoustic rangefinder and data transfer via sockets.}
    \label{fig:sys_diagram}
\end{figure}

In addition to acoustic sensor, we further incorporated a 6 DoF IMU (MPU-6050) on the robot's chassis, an encoder (JGA25-370) to the gearbox driving the drive wheel and a Hall effect sensor (HAL506UA-A) which counts wheel rotations.
The motor and each of these sensors interface with a microcontroller (Arduino MEGA).
A host computer was used to initiate experiments, store the acoustic data from each ranging cycle, and perform data analysis. 
Code on the microcomputer coordinated the trigger and analog sampling events. It further communicates with the microcontroller to control the motor and record encoder and Hall-effect sensor information. A serial communication protocol transferred encoder, IMU, and Hall sensor data between the microcontroller and the microcomputer. 
Then, a communication protocol based on network sockets transferred data between the microcomputer (server node) and the computer (client node). The pipeline for operating the mobile robot and acquiring data from the acoustic sensor is shown in Fig.~\ref{fig:sys_diagram}. 

\section{Experiments}
\label{sec:expts}

We designed three experiments, each related to a different key behavior useful for mobile robotics. In the first experiment, the mobile robot rolled over surfaces with varying texture including hard wood, memory foam, and rugs. This data set was used to demonstrate classification of the robot's rolling surface based on acoustic data. In the second experiment, the mobile robot rolled over small obstacles whose shape may be round or triangular. This data set was used to for obstacle presence and shape classification. In the third experiment, the robot rolled over small, rectangular obstacles and rolls into (but not over) insurmountable obstacles. Here, a heuristic based on first principles measured the height of tall (insurmountable) and short (surmountable) obstacles. In the classification tasks, classifiers trained with IMU data are compared to those trained with acoustic data from the novel sensor.

\begin{figure}
    \centering
    \includegraphics{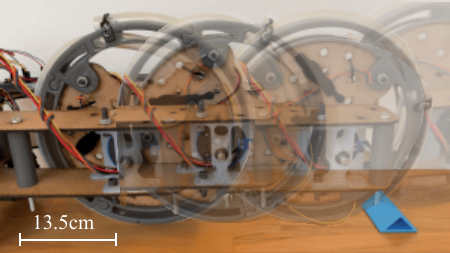}
    \caption{Experimental setup. In this trial, the mobile robot rolls over a triangular obstacle. The opaque image represents the initial position while }
    \label{fig:experiments}
\end{figure}

For each trial of an experiment, we placed the mobile robot on the rolling surface with an initial wheel position $\theta \in \{0\degree, 30\degree, ... 330\degree\ \pm 20\degree\}$ so that throughout all trials, evenly spaced initial positions were sampled. Each experiment began with a sequence of ten seconds during which several ranging cycles were executed and the mobile robot was stationary. Following this initial sequence, the mobile robot rolled into the test environment. We used a PID controller to set the angular velocity of the sensorized front drive wheel was set at 6~RPM in all trials. An example of one experiment trial is shown in Figure \ref{fig:experiments}. Throughout each trial, the IMU and rangefinder were queried and logged data at 20~Hz. Per query, the IMU gave six measurements corresponding to angular and linear accelerations. The acoustic rangefinder gave 4000 acoustic amplitude measurements at a rate of 200~kHz. We further queried and logged the motor encoder (which measured motor rotation) and a Hall effect sensor (which logged a high value at every complete rotation of the wheel) at the same rate. 
\label{sec:exp:data_proc}

We processed the data resulting from each experiment. We observed the raw acoustic signal data for multiple cycles in several pilot experiments over the flat, wooden table. Visible return peaks occurred within the first 2000 amplitude measurements after the send pulse. There was a variable delay of up to $7.5$~ms between initiation of the query and initiation of the send pulse. For this reason, the ranging data of each cycle was shifted to count the beginning of the send peak as \textit{ranging time} $= 0$~ms. Length of the processed ranging cycle was set to 2000 measurements; longer traces were truncated while shorter traces were padded with the average amplitude of all traces in the trial. IMU data for linear acceleration was returned such that for a measurement $x$ where $x/16384$ gives units of $g$'s. IMU data for rotation were such that for a measurement $r$, $r/100$ gives units of degrees. Acoustic amplitude was between 0-1.8~V and read as an ADC value between 0 and 4096. Each channel of the IMU data was baselined by subtracting the mean of a subset of the first measurements. Hall effect sensor and encoder data were recorded with initiation of each ranging cycle.

We split experimental data from each trial into windows. A window of width $N$ contains $N$ truncated acoustic traces (side-by-side), each of length 2000, $N$ 6-DoF IMU measurements, $N$ Hall effect sensor measurements and $N$ encoder measurements. Each of the $N$ measurements was timestamped to the initiation of the send pulse. We describe \textit{experiment time} $t_{ex}$ as the time that a query was initiated within one trial, and \textit{ranging time} $t_r$ as the within one ranging cycle, relative to the initiation of the send pulse, that an amplitude measurement was received. Experiment time was logged at the initiation of a query. Ranging time was not logged directly. Rather, a maximum of 4000 amplitude measurements were gathered per query; enumerating the measurements enables ranging estimation time within a few milliseconds. Times were measured in milliseconds (ms).
 
The sections below explain specific test conditions of each of the three experiments performed and detail any additional data processing and analysis specific to those experiments.

\subsection{Rolling Surfaces with Variable Material} 

In the first experiment, the mobile rolled over each of five surfaces of different material. The five surfaces tested are a wooden table (`Wood'), a textured doormat (`Outdoor'), a memory foam mat (`Soft'), a memory foam mat with seams forming evenly spaced indents (`Ribbed'), and a small rug (`NFM'). Photos are shown in Appendix \ref{appendix:experiment}. The robot began on the wooden rolling surface with the surface material $\approx$0.5 meters from the initial contact position of the wheel. Observing the robot, we used a keyboard flag on the host computer to identify the beginning of the wheel's contact with the material. ``Wood'' did not have a keyboard flag because there was no change of surface material throughout rolling. For each material, we recorded 10 trials, resulting in 50 trials.

\begin{figure}
    \centering
    \begin{subfigure}{0.48\linewidth}
        \centering
        \rotatebox{-90}{\includegraphics[width=2.5\linewidth]{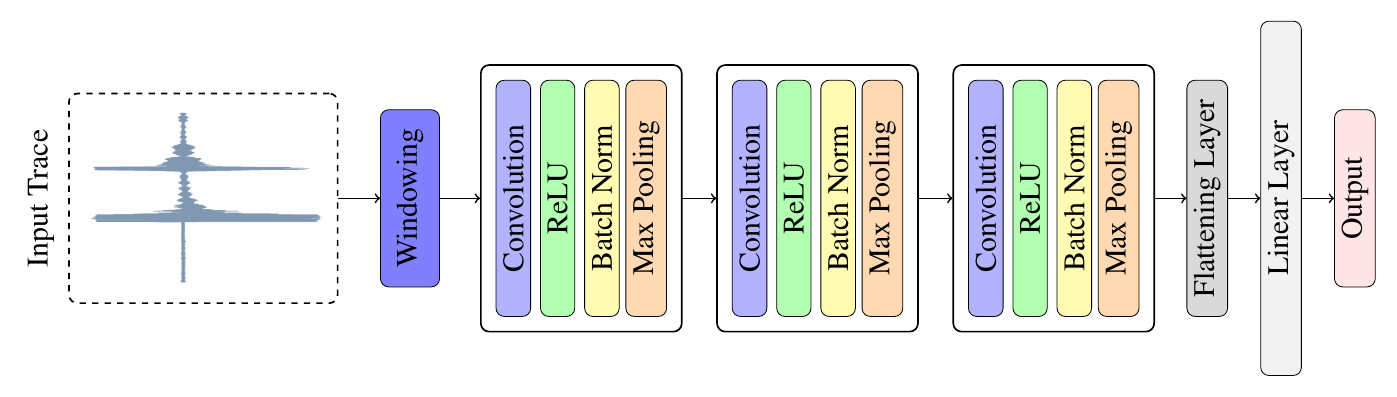}}
        \caption{}
        \label{fig:terrain_CNN_arch_1D}
    \end{subfigure}
    \begin{subfigure}{0.48\linewidth}
        \centering
        \rotatebox{-90}{\includegraphics[width=2.15\linewidth]{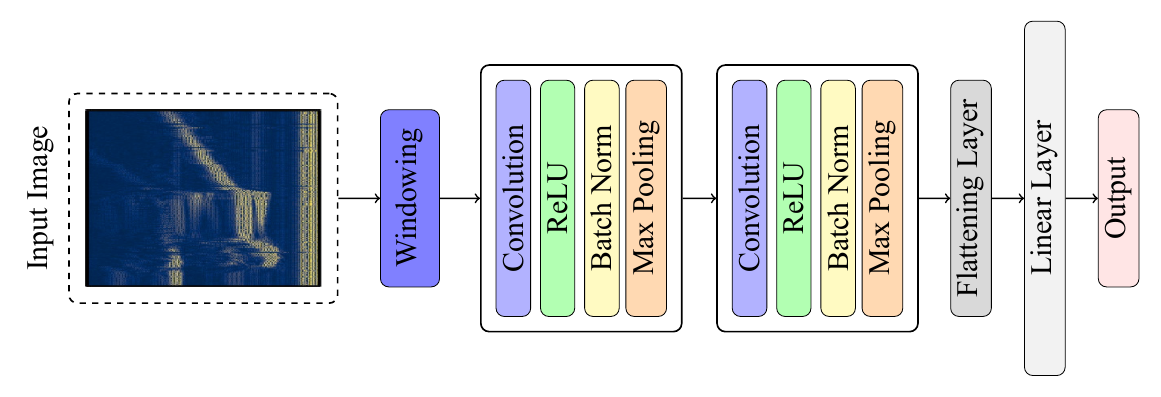}}
        \caption{}
        \label{fig:terrain_CNN_arch_2D}
    \end{subfigure}
    \caption{CNN architectures with acoustic input data in a window of $N$ ranging cycles: (a) 1-D CNN with three convolutional layers. Input here is a vector amplitude data across $t_{r,i}, i \in \{1, 2, ... N\}$. (b) 2-D CNN with two convolutional layers. Acoustic input here (as pictured) is a 2D tensor of amplitude data across $t_{r,i}, i \in \{1, 2, ..., N\}$. IMU input is a 2D tensor of 6-DoF acceleration measurements of size $N \times 6$.}
    \label{fig:terrain_CNN_arch}
\end{figure}
\par 

Rolling surface materials were classified via one- and two-dimensional CNN's. The 1-D CNN underwent minimal hyperparameter tuning, while the 2-D CNN was tuned via an exhaustive grid search. Figure \ref{fig:terrain_CNN_arch_1D} shows the structure of the 1-D CNN. Figure \ref{fig:terrain_CNN_arch_2D} shows the structure of the 2-D CNN. The 1-D CNN had three convolutional layers, one hidden layer with a size of 128, and five outputs for the five classes. The 2-D CNN had a similar structure, but with only two convolutional layers. Hyperparameter ranges searched for the 2-D CNN are visible in Table~\ref{tab:CNN2Dhyperparameters}. 

\begin{table}[h]
    \centering
    \caption{Hyperparameter ranges for 2-D CNN in rolling surface classification. The values related to kernels represent the dimension of one side of a square kernel. The neural net size represents the size of the hidden linear layer.}
    \begin{tabular}{|c|c|}
        \hline
        \textbf{Hyperparameters} & \textbf{Values} \\
        \hline
        Window Size & 2-6 \\
        Epochs (fixed) & 20 \\
        Conv. Kernel 1 Size (C1k) & 2-6 \\
        Conv. Kernel 2 Size (C2k) & 2-6 \\
        MaxPool Kernel 1 Size (mpl1) & 2-6 \\
        MaxPool Kernel 2 Size (mpl2) & 2-6 \\
        Neural Net Size (fixed) & 128 \\
        \hline
    \end{tabular}
    \label{tab:CNN2Dhyperparameters}
\end{table}

To evaluate the effectiveness of each permutation of hyperparameters, three random train-test splits were assessed. Their accuracy and precision were recorded. The best performing CNN (first in overall accuracy and fourth in overall precision) was selected, and so these hyperparameters were chosen for the 2-D CNN: Window Size: 5, Epochs: 20, C1k: 4, C2k: 6, mpl1: 3, mpl2: 3, Neural Net Size: 128.  We classified terrains (i.e., surface material) with the 1-D and 2-D CNNs using acoustic data.
We further applied the same 2-D CNN architecture to the IMU data gathered over the same intervals as the ranging data with the same window size.

\subsection{Obstacles with Variable Shape}
\label{sec:expts:phys:obs} 

In the second experiment, the mobile robot rolled over each of a set of eight small obstacles. Each 3D printed (PLA on Prusa MK3) obstacle had a shape (triangular or round) and one of four possible heights. The full sample space of obstacles is given in Table \ref{tab:obstacles}. Photos are shown in Appendix \ref{appendix:experiment}. The robot began on the wooden rolling surface with the obstacle taped less than one meter from the initial contact position of the wheel. Observing the robot, we used a keyboard flag on the host computer to identify the beginning and end of the wheel's contact with the obstacle. We conducted the experiments in two blocks. In the first block, we exclusively acquired training data from the 25 mm-high obstacles. This resulted in 134 trials per shape. Experiments with the round obstacle and the triangle we alternated to avoid encoding artifacts related to the order of data acquisition. In the second block, we acquired between 15 and 42 trials for each obstacle and size combination.

\begin{table}[]
    \caption{Labels associated with the obstacle detected and classification experiment.}
    \centering
    \renewcommand{\arraystretch}{1.5}
    \begin{tabular}{|c|c|}
        \hline
        \textbf{Task}               & \textbf{Labels}                      \\ \hline
        \textbf{Obstacle Flag} & $\mathcal{D}_1 =$\{0, 1\}                             \\ \hline
        \textbf{Obstacle Shape}     & $\mathcal{D}_2 =$\{Semi-Circle, Triangle \} \\ \hline
        \textbf{Obstacle Height}    & $\mathcal{D}_3 =$\{10, 15, 20, 25\} [mm]                  \\ \hline
    \end{tabular}
    \label{tab:obstacles}
\end{table}

We assessed three supervised methods to classify the obstacles. The first was a logistic regression model and the next two were 2-D CNNs. We performed additional pre-processing to the procedure described in Sec. \ref{sec:expts}. Acoustic traces were baselined by subtracting data that had undergone a low-pass filter ($f_{\mathrm{cutoff}}=100 \mathrm{Hz}$) from the raw signal. The baselined signal was then rectified by taking its absolute value. Each baselined and rectified trace was shifted to align the send pulses. Each trace was independently normalized by the maximum amplitude within it. We truncated the signal by eliminating the first 350 elements (i.e., the send pulse) and by only retaining 1750 samples thereafter. The inputs to the models were windows of labelled data extracted from each experiment. The size of each window was $90 ~[\mathrm{windows}] \times 1750~ [\mathrm{samples}]$. From each trial, we extracted exactly one window corresponding to the obstacle. In the obstacle window, a maximum of 15 traces could be included before the first keyboard flag; data after the second keyboard flag were not restricted. Obstacle windows could not be extracted from some trials since the window did not fit within the applicable data; therefore. We extracted flat-ground data from the same trials using data bounded by the first trace and 15 traces from the first keyboard flag. The total number of windows in this data set is then 858.

We trained the logistic regression classifier and the CNNs on windows data from the first experimental block (i.e., obstacles of both shapes, height 25~mm). We tested each classifier on windowed data of the second experimental block (i.e., obstacles of all heights and shapes). The second block contained trials for obstacles of height $\{10, 15, 20, 25~mm \}$ but 25-mm data were new measurements and as such no data was shared between the train and test sets. To train the logistic regression classifier, we flattened the acoustic traces into a single vector of length 157,500 ($90 ~[\mathrm{windows}] \times 1750~ [\mathrm{samples}]$) samples. We used the logistic regression function from Scikit-learn with L2 regularization and an inverse regularization strength ($C$) of 0.5.

We constructed CNNs in PyTorch based on two convolutional blocks. Each block contained a two-dimensional convolutional layer, an activation layer, a batch normalization layer, and a max pooling layer. The output of the second convolutional block is flattened and passed into a fully connected layer. The three-element output vector of the fully connected layer is used to predict the class of the input window.

\begin{table}[h]
    \centering
    \caption{Hyperparameters of CNN used for obstacle presence and shape classification.}
    \begin{tabular}{|c|c|}
\hline
\textbf{Hyperparameters}     & \textbf{Value} \\ \hline
Window Size                  & 90               \\
Epochs                       & 30               \\
Conv. Kernel 1 Size (C1k)         & 8                \\
Conv. Kernel 2 Size (C2k)         & 4                \\
MaxPool Kernel 1 Size (mpl1)      & 3                \\
MaxPool Kernel 2 Size (mpl2)       & 2                \\
Fully-Connected Layer Length & 64               \\ \hline
\end{tabular}
    
    \label{tab:obs_CNN2Dhyperparameters}
\end{table}

\begin{figure}
    \centering
    \begin{subfigure}{0.15\textwidth}
        \includegraphics[width=0.9\textwidth]{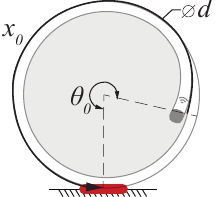}
        \caption{Ground contact at $t_{ex,0}$.}
        \label{fig:wheel_heuristc}
    \end{subfigure} \hfill
    \begin{subfigure}{0.15\textwidth}
        \includegraphics[width=\textwidth]{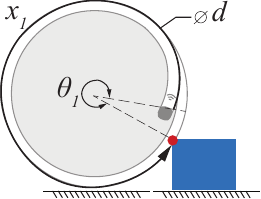}
        \caption{Obstacle contact at $t_{ex,1}$.}
        \label{fig:wheel_heuristic_obs}
    \end{subfigure} \hfill
    \begin{subfigure}{0.15\textwidth}
        \includegraphics[width=\textwidth]{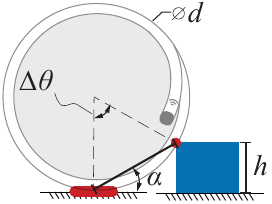}
        \caption{Ground and obstacle contact.}
        \label{fig:wheel_heuristic_obs}
    \end{subfigure}\\
    \vspace{6pt}
    \begin{subfigure}{0.5\textwidth}
        \includegraphics[width=\textwidth]{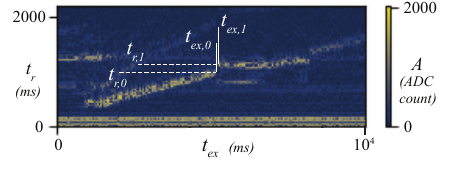}
        \caption{Acoustic trace with obstacle.}
        \label{fig:wheel_heuristic_obs_2}
    \end{subfigure}    
    \caption{Change of contact of wheel with outer diameter $d$ while surmounting an obstacle. (a) Ground contact at experiment time $t_{ex,0}$ with angle $\theta_0$ and distance $x_0$ from rangefinder, (b) rectangular obstacle contact at experiment time $t_{ex,1}$ with angle $\theta_1$ and distance $x_1$ from rangefinder (c) ground and obstacle contact (d) example of acoustic trace of wheel rolling over obstacle with $t_{r,0}$ as the ranging time to the return peak at experiment time $t_{ex,0}$ and $t_{r,1}$ as the ranging time to the return peak at experiment time $t_{ex,1}$.}
    \label{fig:obs_height}
\end{figure}

\subsection{Contact Location and Variable Obstacle Height} 

In the third experiment, the robot rolled into each of two rectangular obstacles of different sizes. The smaller obstacle was short and surmountable (height = 2.5 cm) whereas the larger obstacle was tall and insurmountable (height = 7 cm). Photos are shown in Appendix \ref{appendix:experiment}. We began each trial by placing the mobile robot on a wooden table. One of the two rectangular obstacles was placed within one meter of the initial contact location of the front wheel of the mobile robot. The robot then rolled into the obstacle, climbing the surmountable obstacle and remaining stationary after colliding with the insurmountable obstacle. We used a keyboard flag to manually identify the beginning of wheel contact with the obstacles. We acquired acoustic peak data throughout each trial and used a keyboard flag to indicate initiation of contact with the obstacle. We acquired 12 trials with the tall obstacle and 43 trials with the small obstacle.

We built the heuristic by observing that collision with an obstacle cause an abrupt change in the wheel's contact location and therefore in the timing of $t_{r}$ of the reflected peak. The reflected peak ranging time in $t_{r}$ was be used to determine the location along the wheel's perimeter of the contact. Figure \ref{fig:obs_height} shows this phenomenon. Parts (a), (b) and (c) show the transition from floor to obstacle contact, and (d) shows an example of such an event within an acoustic sensor measurement. In the case of transitioning from ground contact at $t_{ex,0}$ to obstacle contact at $t_{ex,1}$, the difference in peak location $t_{r,1}-t_{r,0}$ gives information about the obstacle's height. We used first principles to find the angular location of contact of the waveguide on the wheel. We then compared the locations of contact between ranging cycles to measure obstacle height. A few assumptions guide this analysis. We used a simplified model of a rigid wheel rolling without slippage. There are up to 2 contact points per ranging cycle (e.g., Fig. \ref{fig:wheel_heuristic_obs_2}a-c) and both create acoustic return peaks. Ranging cycles occur at exactly 20Hz and amplitudes are read in at $200$kHz without fluctuations. Though the rangefinder overlaps with the interior of the wheel, it is treated as though it perfectly follows the perimeter. Equation \ref{eq:rolling_contact_dist} gives the relationship between wheel outer diameter $d$, angle between rangefinder and contact $\theta$, and distance between rangefinder and contact location along the wheel's perimeter $x$:

\begin{equation}
    x = \theta \frac{d}{2}.
    \label{eq:rolling_contact_dist}
\end{equation}

Figure \ref{fig:wheel_heuristic_obs_2} shows the geometric relationship between contact points and contact locations $\theta_1 - \theta_0 = \Delta \theta$ along the wheel's perimeter. The sound travels a distance of $2x$ to go from the rangefinder to the obstacle off of which it is reflected, and back. The difference in return peak time before and after collision is defined as $\Delta t_{c} = t_{r,1} - t_{r,0}$. Assuming sound travels at speed $c$ and using Eqn. \ref{eq:rolling_contact_dist}, $\Delta t_c$ and $\Delta \theta$ are related by:

\begin{equation}
    \Delta\theta = \frac{c \Delta t_{c}}{d}.
    \label{eq:time_of_flight}
\end{equation}

An obstacle of height $h$ can be written in terms of angle $\alpha$ and the change in contact location $\Delta \theta$ as follows:

\begin{equation}
    \sin(\alpha) = \frac{h}{d  \sin\big(\frac{\Delta \theta}{2}\big)}. \\
\end{equation}

Noting that, because the triangle containing the chord is isoceles, $\alpha = \frac{\Delta \theta}{2}$,

\begin{equation}
    h = d \sin^2 \big( \frac{\Delta \theta}{2} \big).
    \label{eq:obs_height}
\end{equation}

Equation \ref{eq:time_of_flight} relates times of peaks in $t_r$ on each acoustic trace to angular wheel positions. 
To isolate the peaks, we scaled (normalized by the highest amplitude value in the trace) and filtered each acoustic trace with an exponential moving average (EMA) filter. SciPy's ``find peaks'' function isolated the peaks.Details of parameter selection and an example of peak data are given in Appendix \ref{appendix:peaks}. The quantity $\Delta \theta$ was constrained to be positive and less than $2\pi$. We performed peak analysis on the obstacle types testing in the experiment corresponding to this task. In each trace, the keyboard flag gave the experiment time $t_{ex,c}$ corresponding to obstacle collision. We used a window of $\pm \varepsilon = 500ms$ around $t_{ex,c}$ and measured $\Delta t_{c}$ as the difference between the 80th percentile and 20th percentile return peak distances. Eqns. \ref{eq:time_of_flight} and \ref{eq:obs_height} converted measurements of $\Delta t_c$ to obstacle heights.

\section{Results}
\label{sec:results}

	\vfill
\begin{figure*}
    \begin{subfigure}{0.33\textwidth}
        \includegraphics[width=\textwidth]{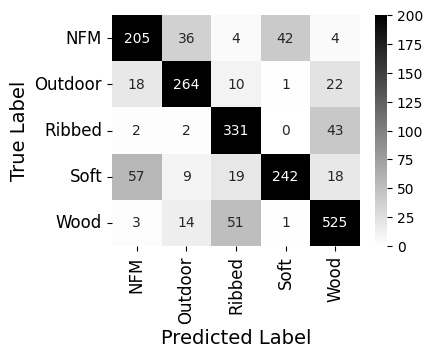}
        \caption{}
        \label{fig:confusion_terrain_acoustic1D}
    \end{subfigure} \hfill
    \begin{subfigure}{0.33\textwidth}
        \includegraphics[width=\textwidth]{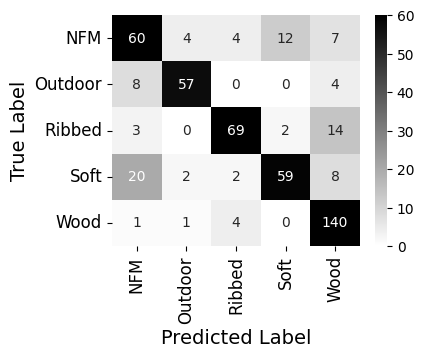}
        \caption{}
        \label{fig:confusion_terrain_acoustic2D}
    \end{subfigure} \hfill
    \begin{subfigure}{0.33\textwidth}
        \includegraphics[width=\textwidth]{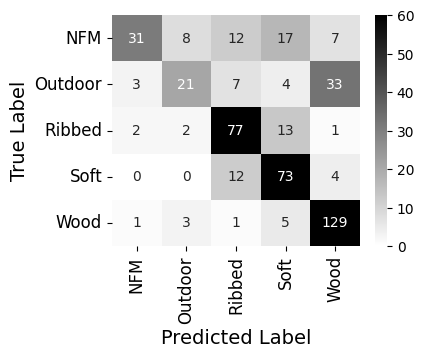}
        \caption{}
        \label{fig:confusion_terrain_imu}
    \end{subfigure} \\

    \vfill
    
    \begin{subfigure}{0.33\textwidth}
        \includegraphics[width=\textwidth]{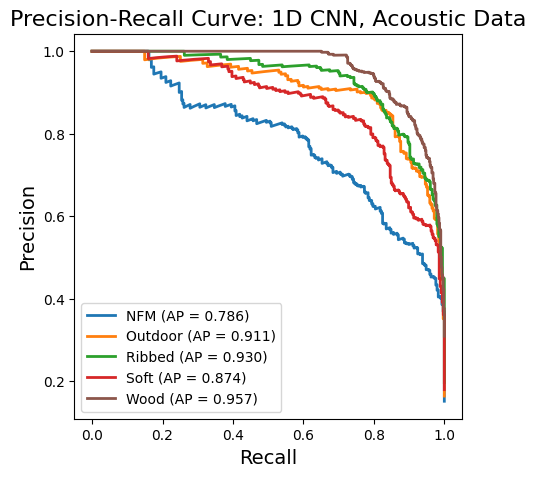}
        \caption{}
        \label{fig:prc_terrain_acoustic1D}
    \end{subfigure} \hfill
    \begin{subfigure}{0.33\textwidth}
        \includegraphics[width=\textwidth]{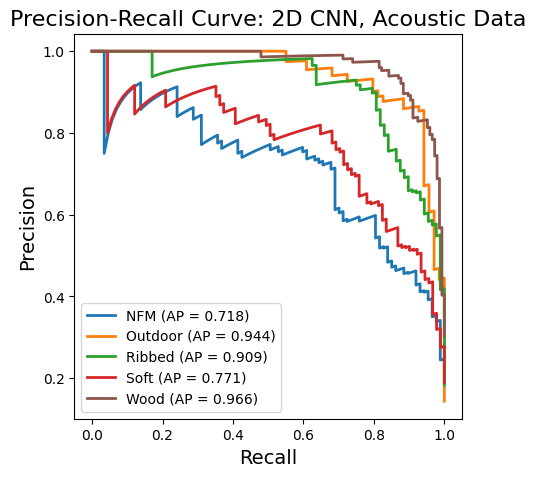}
        \caption{}
        \label{fig:prc_terrain_acoustic2D}
    \end{subfigure} \hfill
    \begin{subfigure}{0.31\textwidth}
        \includegraphics[width=\textwidth]{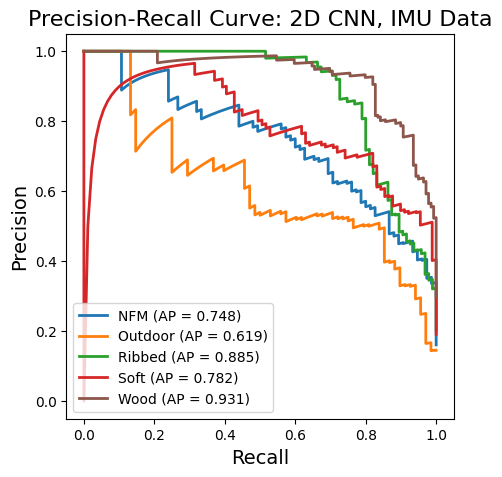}
        \caption{}
        \label{fig:prc_terrain_imu}
    \end{subfigure}
    
    \caption{Confusion matrices and Precision-Recall Curves (PRC): (a) Confusion matrix and (d) PRC for 1-D CNN trained on acoustic ranging data; (b) Confusion matrix and (e) PRC for 2-D CNN trained on acoustic data; (c) Confusion matrix and (f) PRC for 2-D CNN trained on IMU data.}
    \label{fig:terrain_metrics}
\end{figure*}

\begin{figure}
    \begin{subfigure}{0.225\textwidth}
        \includegraphics[width=\textwidth]{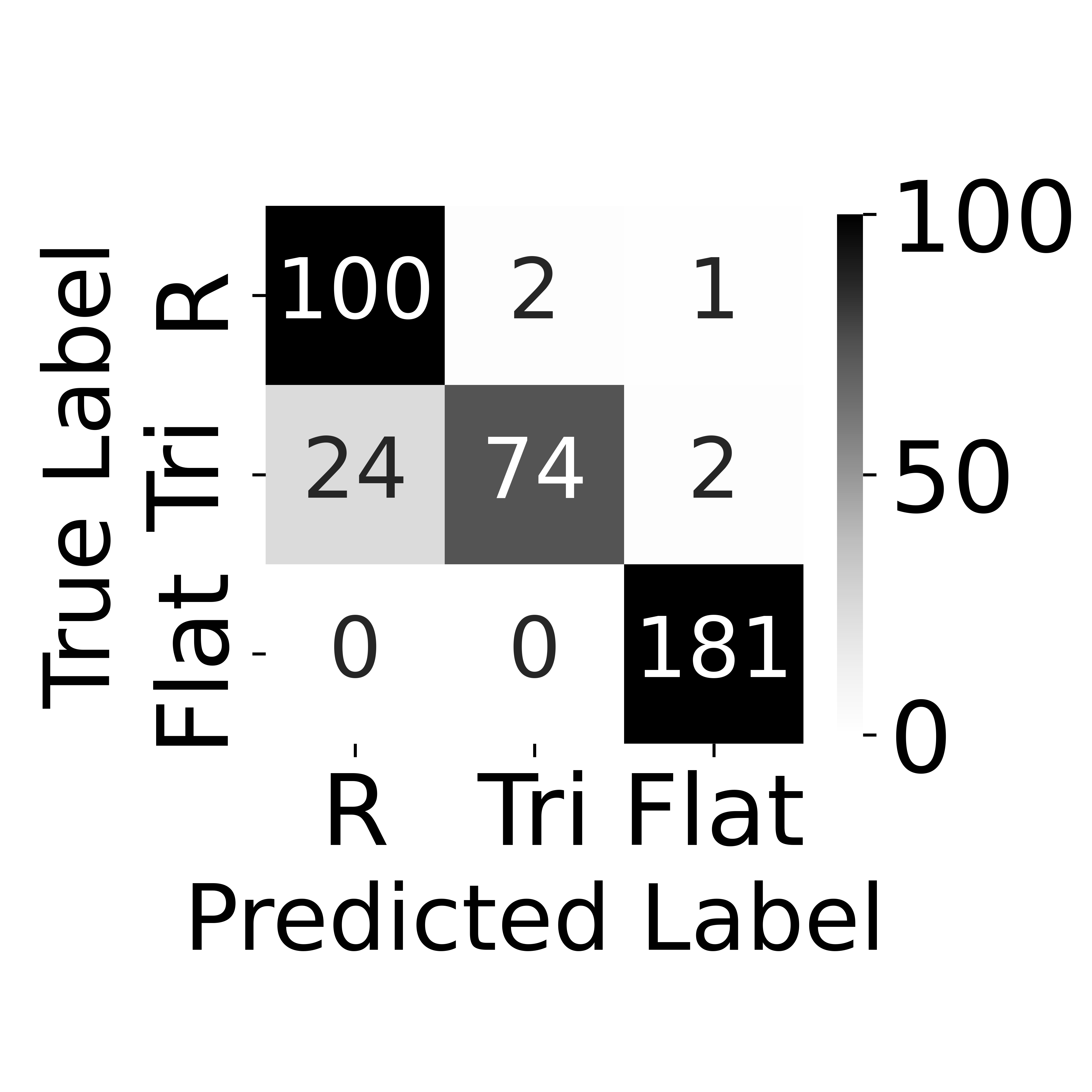}
        \caption{Results from 2-D CNN trained on acoustic data.}
        \label{sfig:confusion_obs_acoustic}
    \end{subfigure} \hfill
    \begin{subfigure}{0.225\textwidth}
        \includegraphics[width=\textwidth]{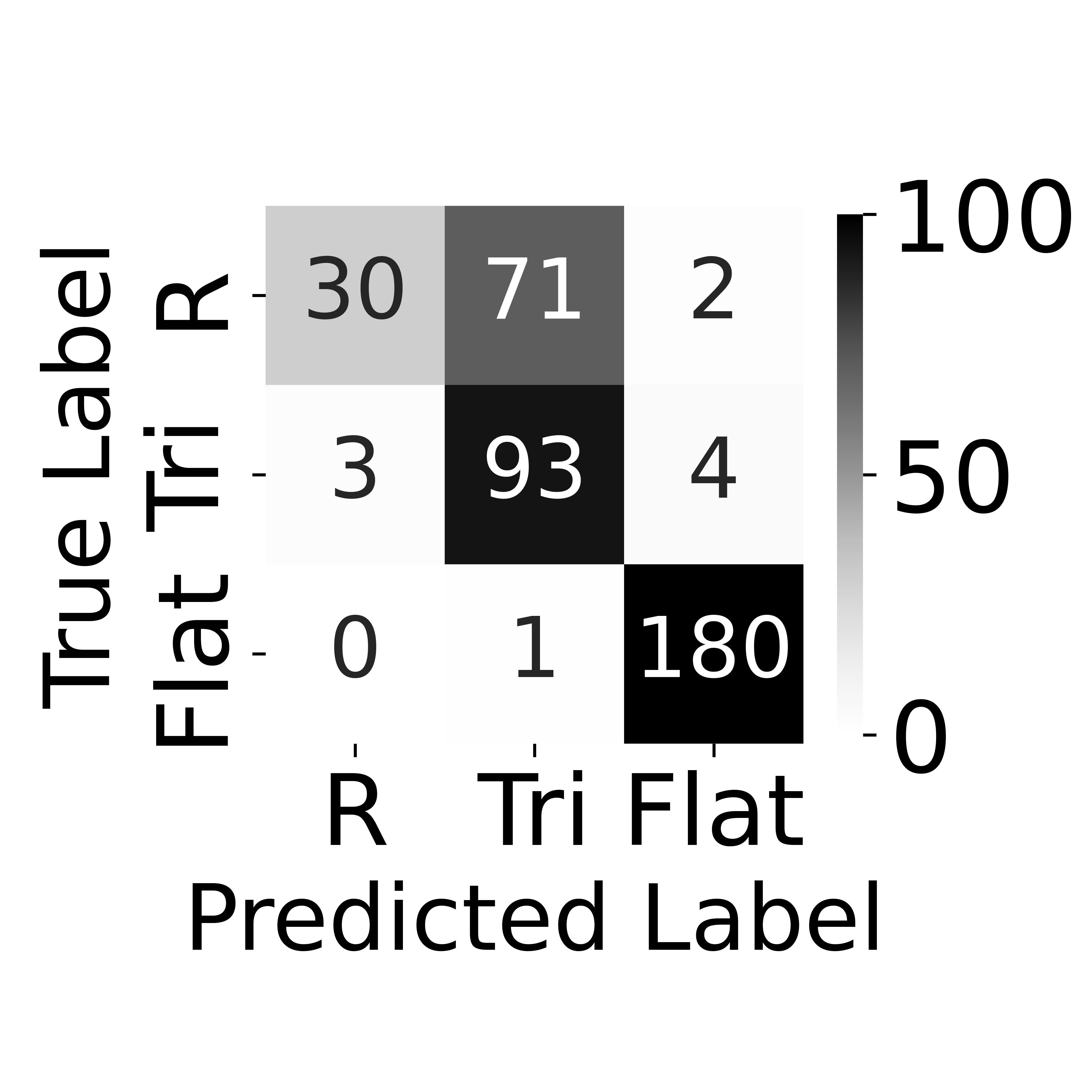}
        \caption{Results from the 2-D CNN trained on IMU data.}
        \label{sfig:confusion_obs_imu}
    \end{subfigure}
    \caption{Confusion matrices for the highest accuracy models trained on either acoustic ranging data or IMU data (R: round obstacle, Tri: triangular obstacle, Flat: flat-ground)}
    \label{fig:confusion_obs}
\end{figure}

We evaluated accuracy and precision of rolling surface material classifiers trained/tested on acoustic data in comparison with the same classifier architectures trained/tested with IMU data. We similarly evaluated obstacle presence and shape classifiers trained on acoustic and IMU data. In the third experiment, we evaluated the obstacle height measurements output by the heuristic.

\subsection{Rolling Surface Classification}
The 1-D CNN trained and tested on acoustic ranging data achieved the best results of the methods explored. The 2-D CNN trained on acoustic ranging followed. The 2-D CNN trained on tested on IMU data had success but the lowesr accuracy and precision among the three. The 1-D CNN outperformed the 2-D CNN on average despite minimal hyperparameter tuning. 

\begin{table}
    \centering
    \caption{Test Accuracy and Precision for 1-D and 2-D CNN architectures in rolling surface material (terrain) classification, averaged over five random train-test splits, each trained either on acoustic or IMU data.}
    \begin{tabular}{|c|c|c|}
        \hline
        \textbf{Architecture (Data Type)} & \textbf{Accuracy [\%]} & \textbf{Precision [\%]} \\
        \hline
        2-D CNN  (Acoustic Ranging) & 77.63 & 77.41 \\
        1-D CNN (Acoustic Ranging) & \textbf{78.38} & \textbf{79.99} \\
        2-D CNN (IMU) Data & 74.94 & 73.52 \\
        \hline
    \end{tabular}
    \label{tab:performance_metrics}
    
\end{table}

Figures \ref{fig:terrain_metrics}a-c show that the 1-D CNN trained on acoustic data had the best performance on a per-class basis, followed by the 2-D CNN. The 1-D CNN was trained and tested on approximately five times the amount of data as the 2-D CNNs, due to the use of non-overlapping windows of width five. For this reason, \ref{fig:confusion_terrain_acoustic1D} has more data points and a corresponding scale adjustment to accurately represent the data. The two acoustic classifiers performed the worst on the ``NFM" category, while the IMU classifier performed the worst in the ``Outdoor" category and second-worst in the ``NFM'' category. Figure \ref{fig:terrain_metrics}(d)-(f) shows precision-recall curves separated by material for each classifier. Classifiers trained with acoustic data have the lowest precision-recall for the ``NFM'' category (an area rug that is softer than the door mat but firmer than the other mats), while the IMU-trained classifier performs worst on ``Outdoor''. All classifiers perform best on ``Wood'' in terms of precision and recall.

\subsection{Obstacle Presence and Shape Classification}

\begin{table}
\centering
\caption{Accuracy and precision of logistic regression (LR) and two-dimensional convolutional neural network (2-D CNN) models trained and tested on acoustic ranging or inertial measurement unit (IMU) data. The results for the 2-D CNN models are reported as the average across 14 random train-test splits plus or minus one standard deviation.}
\label{tab:obstacle_results}
    \begin{tabular}{|c|c|c|}
        \hline
        \textbf{Architecture (Data Type)} & \textbf{Accuracy [\%]} & \textbf{Precision [\%]} \\
        \hline
        2-D CNN (Acoustic Ranging Data) & \textbf{89.02 $\pm$ 2.45} & \textbf{88.76 $\pm$ 3.03} \\
        LR (Acoustic Ranging Data) & 82.55 & 78.95 \\
        2-D CNN (IMU Data) & 75.37 $\pm$ 2.20 & 75.20 $\pm$ 2.88 \\
        LR (IMU Data) & 70.05 & 63.25 \\ 
        \hline
    \end{tabular}
\end{table}

Table \ref{tab:obstacle_results} shows the accuracy and precision scores for the two models (2-D CNN and logistic regression) trained and tested on either acoustic ranging data or IMU data. The best-performing models are the 2-D CNN (averaged across 14 seeds), followed by the logistic regression model trained on acoustic ranging data. The models trained exclusively on IMU data achieved the lowest overall accuracy and precision scores.

Fig. \ref{fig:confusion_obs} shows the confusion matrices for the models with the highest accuracy trained on each datatype. The highest accuracy for shape classification was 92.45\% and was achieved with a 2-D CNN trained on acoustic data  (Fig. \ref{sfig:confusion_obs_acoustic}). The lowest class-specific accuracy achieved by this model was 74\% for the triangular obstacle. The highest accuracy for shape classification using IMU data (78.91\%) was also achieved with a 2-D CNN (Fig. \ref{sfig:confusion_obs_imu}). The lowest class-specific accuracy achieved by this model was 68.93\% for the round obstacle. Both the CNN trained on acoustic data and IMU data achieved similar classification accuracies on flat-ground windows.

\subsection{Obstacle Height Measurement}

\begin{figure}
    \centering
        \begin{subfigure}{0.45\textwidth}
        \centering
        \includegraphics[width=0.65\textwidth]{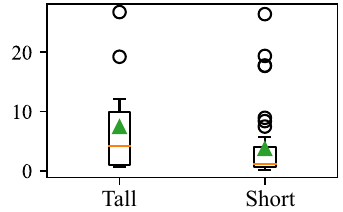}
        \caption{Measured height $h$ across all trials for Tall and Short obstacles. Median as orange line, mean as green triangle, interquartile range (IQR) as box, whiskers as IQR $\pm1.5$~IQR, and outliers as circles.} 
        \label{fig:boxplot_heuristic}
    \end{subfigure} \\
    \vspace{10pt}
    \begin{subfigure}{0.45\textwidth}
        \includegraphics[width=\textwidth]{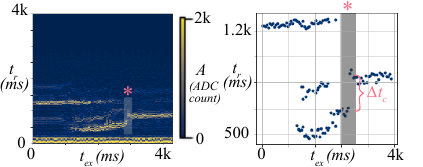}
        \caption{Accurate $\Delta t_c$ measurement through heuristic method. Raw trace (left) with processed peaks (right). Window around collision flag as * in each panel.}
        \label{fig:good_heuristic_trace}
    \end{subfigure}
    \vfill
    \caption{Examples of heuristic usage and obstacle height measurement results across all trials. (a) Box plot showing measured heights $h$ of the tall and short obstacles. (b) Example of an accurate measurement, in which measured $\Delta t_c$ corresponds only to acoustic peaks caused by the collision.}
    \label{fig:heuristic_traces}
\end{figure}

Accuracy of the heuristic depends on existence of peaks unrelated to the obstacle contact. Fig. \ref{fig:good_heuristic_trace} shows an example of use of the heuristic where such peaks do not interfere. 
Figure \ref{fig:boxplot_heuristic} shows the measured height from all 12 trials of the tall obstacle and all 43 trials of the short obstacle. For both the tall and the short obstacle, performance varies widely but the median measurements are within 2.8~cm and 1.4~cm of the true height for the tall and short obstacles respectively. Median $h$ measured for the tall obstacle is 4.1~cm (true height 7~cm). The interquartile range of measurements was 8.9~cm and the total range was 26.1~cm. Median $h$ measured for the short obstacles is 1.1~cm (true height 2.5~cm). The interquartile range is 3.3~cm and the total range is 26.2~cm. High-measurement outliers exist for both the tall and short obstacle. We observed the traces and peaks for all outliers and found that peaks unrelated to the obstacle contact caused this error; for example, the bright peak at $t_r = 20$~ms on Fig. \ref{fig:good_heuristic_trace} would have caused an error had the collision occurred earlier in the measurement sequence. In spite of the large measurement ranges, medians and IQR show the correct broad trend, trending higher for the taller obstacle and lower for the short obstacle.

\section{Discussion}
\label{sec:discussion}

This paper investigates the use of a new acoustic, tactile sensor that functions jointly as the tire of the mobile robot. This acoustic sensing configuration enables continuous contact sensing around the circumference of the mobile robot's wheel in conditions where some camera-based systems might suffer from occlusions and where fiber-optic systems might lack the needed deformability. Acoustic `traces' consisting of send and reflected ultrasound pulses provide high reflectivity where the tire is deformed and varying reflection patterns depending on the tire's contact surface and contact forces. Through a series of three experiments, we demonstrated that the proposed sensor provides acoustic temporal information that can be used in terrain classification, identification of the shape of small features and approximate measurement of obstacle height in collision. Classifiers trained and tested with acoustic data either matched or significantly outperformed those trained with IMU data.

The first task evaluated was terrain classification, between a set of five rolling surface materials. We evaluated the acoustic data in a one-dimensional and two-dimensional convolutional neural network pipelines, achieving peak overall test accuracies of 78.38 \% (1-D CNN) and 77.63 \% (2-D CNN). As a comparison, we used the same classifier with IMU data arising from the same experiment. The 2-D CNN classifier trained with acoustic data has comparable overall performance to the IMU-based classifier. 

Confusion matrices for rolling surface classification show relatively low misclassification rates. Precision-recall curves demonstrate reasonable performance across all classes in spite of imbalanced training data which contained two times more measurements from `Wood' than any other class. The worst performance in the range was observed for the `NFM' and `Outdoor' area rugs which are similar in texture.

The second task evaluated was classification of between flat ground and obstacles with round or triangular shape. We evaluated a logistic regression classifier and a 2-D CNN, each trained and tested with acoustic and IMU data. All classifiers showed the ability to distinguish obstacles from flat ground and some shape classification abilities. Further, the classifiers generalized to new obstacle sizes not present in the training sets. In this task, classifiers trained and tested with acoustic data significantly outperformed those trained with IMU data.

Confusion matrices for obstacle presence and shape classification show that the IMU-trained classifier distinguishes whether the robot is rolling over a flat surface or obstacle, but has a lower success rate classifying between the round or triangular obstacle when an obstacle is present. In contrast, the acoustic-trained classifiers correctly classify the round vs. triangular obstacles at a higher rate. The lowest class-specific accuracy for the 2-D CNN trained on acoustic data is for the triangular obstacle, which is sometimes misclassified as round. It is possible that this error occurs when multiple reflections from a triangular obstacle are weak or obscured in the acoustic traces. Based on visual differences between traces of different shapes (e.g., triangular in Fig. \ref{fig:tri_trace} and round in Fig. \ref{fig:round_trace}), we hypothesize that information including (1) multiple reflections from obstacle contact and (2) variations in the location of obstacle contact traces enable important size-invariant features to be learned. 

For the third task, we developed and evaluated a heuristic that uses acoustic data to measure the heights of tall (7~cm) and short (2.5~cm) rectangular obstacles. Measurements taken with the heuristic had the correct broad trend, with median measurements within 2.8~cm and 1.4~cm of the true obstacle heights. Appearance of peaks unrelated to obstacle contact made the measurements unreliable overall, causing several trials to have high measurement error. We hypothesize that these non-obstacle peaks are caused by resonant phenomena and ambient noise. These causes of error could be addressed with improved perception algorithms. A custom peak-jump function with RANSAC \cite{fischler1981random} could enable a heuristic that ignores irrelevant peaks. Bayes filters that take peaks from previous experiment time steps into account as priors could also improve performance. Further, better physical models of vibrations in the waveguide could form the basis for heuristics that make better use of resonant phenomena. Given trend-wise success of the simple heuristic presented here, the above methods will form the basis of future work that uses the sensor in more complex environments.

The system presented here has a specific wheel morphology; it is a thin wheel in which the wheel radius is more than ten times the tire radius. Other wheels (e.g., cars) may have different geometries. Coiling the sensor with multiple turns could enable use of the tire on wheels with larger ground contact patches. Yet, fundamental dimensional limitations still exist that create trade-off between sensor speed and length, as well as coiling diameter and the acoustic wavelength required. Future work would seek to adapt this sensor to a wider array of wheel morphologies and empirically characterize these trade-offs. The work here nonetheless shows that using the waveguide jointly as the tire and part of the sensor enables full-perimeter tactile sensing.

The experiments' spatial setup and set window size caused training data set imbalance. They included higher proportions of `Wood' in the terrain classification task and flat ground in the obstacle classification task as this was the initial state of the robot before reaching the rugs or obstacles. `Wood' and flat-ground windows predominantly come from the first 90 traces ($\approx 45\%$ of the exposed tube length), which often have higher reflected amplitude. Future work will use variable-length sliding windows to provide a more balanced training set and enable classification close to real-time.

In this paper, classifiers trained on acoustic data performed with comparable or higher accuracy and precision than the same classifiers trained on acoustic data. This result provides evidence that the proposed sensor provides information useful to vehicles. Yet, an exhaustive comparison is impossible: in-situ sensor comparisons will always require specific choices related to data processing and classifier architecture to accommodate differences in measurement data type and dimension. It is not possible to exhaustively characterize every possible data filtering technique, classification, and hyperparameter combination. Other pipelines not investigated here might enable better IMU performance. We nonetheless believe that the experiments and classifiers presented here provide reasonable benchmarks and framework choices to validate and confirm the usefulness of the acoustic sensor.

Together, the experiments presented here constitute over 500 trials (and over 1km of rolling) of the sensorized wheel. Wear and tear on the waveguide have not yet been observed. In spite of this promise, a study would be needed to determine the lifespan of this system.

These results indicate that the acoustic sensor has advantages in terrain identification and shape classification on a per-window basis. The heuristic demonstrates the advantage of the sensor to perform measurements anywhere along the circumference of the wheel without multiple speakers or microphones. In comparison with an IMU, the acoustic sensor has a fundamental bandwidth limitations related to the speed of sound and practical bandwidth limitation due to the need for thousands of individual data points in contrast with the IMU's six accelerations. Future pipeline improvements and more sophisticated sensor fusion techniques between the sensor and existing vehicle sensors could yield improvements across mobile robots. The work presented here shows that incorporating acoustic-tactile sensing could be added to mobile robot perception pipelines to yield important information about the robot's ground contact and the terrain itself.

\section{Conclusion}
\label{sec:conclusion}
This paper presents an acoustic, tactile sensor that jointly functions as the tire of a mobile robot. This sensor advantageously integrates a single time-of-flight acoustic rangefinder with a deformable, flexible tube waveguide to create a tactile sensor that conforms around the circumference of the wheel without the need for sensor arrays. We characterized the theoretical limits of this sensor and explained the working premise that acoustic amplitude measurements taken over time can provide useful tactile data. We built a prototype of the sensor and mounted it to a custom mobile robot. We provided a suite of experiments that demonstrate the sensor's capabilities in terrain classification, small object shape classification, and collision localization.

Given the demonstrated capabilities, fuller integration into autonomous mobile robots is planned. The data provided by this sensor could be used in fine-grained mapping tasks, classification of terrain material or underlying structure, and, when mounted to each wheel, reliable contact detection and monitoring for each tire. Future capabilities may also include slip detection, estimation related to vehicle states such as weight balance, and other proprio- and exteroceptive quantities. The design, integration, and suite of experiments given here provide the basic, crucial knowledge needed for roboticists to deploy this sensor.

\section{Dataset and Code}
The dataset will be made publicly available at \url{http://macrobotics.cim.mcgill.ca/datasets}. The code will be made publicly available in our research group's Github organization.

\section*{Acknowledgment}

This research was funded by Meta Reality Labs and NSERC Discovery Grants. The authors would like to thank Lucas Berry for advice on CNN implementation and Madison Odabassian for helping with data acquisition.

\ifCLASSOPTIONcaptionsoff
\newpage
\fi

\bibliographystyle{IEEEtran}
\bibliography{bibtex/bib/IEEEabrv, bibtex/bib/IEEEexample}

\pagebreak
\appendices

\section{Experimental Details}
\label{appendix:experiment}

This appendix contains photos of all obstacles and terrains used in the experiments presented.

\begin{figure}[h]
    \centering
    \includegraphics{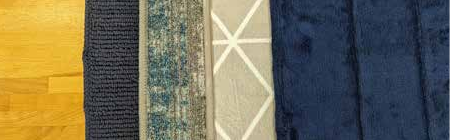}
    \caption{Photo of all rolling surfaces tested. From left: wooden table (`Wood'), rug (`NFM'), door mat (`Outdoor'), memory foam mat (`Soft'), and memory foam mat with seams (`Ribbed').}
    \label{fig:terrain_photo}
\end{figure}

\begin{figure}[h]
    \centering
    \includegraphics{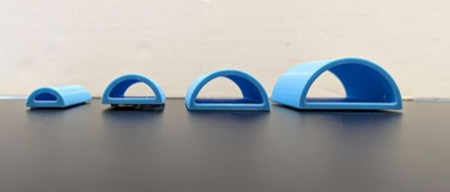}
    \caption{Round (semi-circular) obstacles. Height, from left: 10mm, 15mm, 20mm, 25mm.}
    \label{fig:tri_obs_photo}
\end{figure}

\begin{figure}[h]
    \centering
    \includegraphics{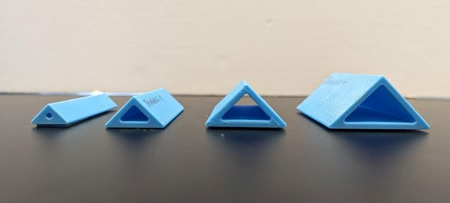}
    \caption{Triangular obstacles. Height, from left: 10mm, 15mm, 20mm, 25mm.}
    \label{fig:tri_obs_photo}
\end{figure}

\begin{figure}[h]
    \centering
    \includegraphics{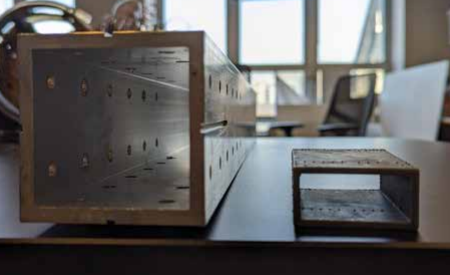}
    \caption{Tall (insurmountable) and short (surmountable) obstacles. Height, from left: 70mm, 25mm.}
    \label{fig:tri_obs_photo}
\end{figure}

\section{Peak Isolation}
\label{appendix:peaks}
\begin{table}[h]
    \centering
    \caption{Parameters used to isolate acoustic peaks in obstacle height heuristic.}
    \begin{tabular}{|c|c|}
    \hline
    Parameter & Value \\
    \hline
    $\alpha_{\mathrm{EMA}}$ & 0.75 \\
    \hline
    $\texttt{height}$  & 0.3 \\
    \hline
    $\texttt{distance}$ & 20 \\
    \hline
    $\texttt{prominence}$ & 0.6 \\
    \hline
    $\texttt{threshold}$ & 0.0001 \\
    \hline
    \end{tabular}
    \label{tab:peak_params}
\end{table}

\end{document}